\pdfoutput=1

\documentclass[11pt]{article}

\usepackage[]{acl}

\usepackage{times}
\usepackage{latexsym}
\usepackage{adjustbox}
\usepackage{pifont}
\usepackage{multirow}
\usepackage{hyperref}
\usepackage{hyperref}
\makeatletter
\def\UrlAlphabet{%
    \do\a\do\b\do\c\do\d\do\e\do\f\do\g\do\h\do\i\do\j%
      \do\k\do\l\do\m\do\n\do\o\do\p\do\q\do\r\do\s\do\t%
      \do\u\do\v\do\w\do\x\do\y\do\z\do\A\do\B\do\C\do\D%
      \do\E\do\F\do\G\do\H\do\I\do\J\do\K\do\L\do\M\do\N%
    \do\O\do\P\do\Q\do\R\do\S\do\T\do\U\do\V\do\W\do\X%
      \do\Y\do\Z}
\def\UrlDigits{\do\1\do\2\do\3\do\4\do\5\do\6\do\7\do\8\do\9\do\0}
\g@addto@macro{\UrlBreaks}{\UrlOrds}
\g@addto@macro{\UrlBreaks}{\UrlAlphabet}
\g@addto@macro{\UrlBreaks}{\UrlDigits}
\usepackage{threeparttable}
\usepackage{booktabs}
\usepackage[T1]{fontenc}

\usepackage[utf8]{inputenc}
\usepackage{CJKutf8}

\usepackage{microtype}

\title{Distinguish Before Answer: Generating Contrastive Explanation as Knowledge for Commonsense Question Answering}

\author{
Qianglong Chen\textsuperscript{1,2}\thanks{\quad Work is done during internship at Alibaba Group.}, Guohai Xu\textsuperscript{2}, Ming Yan\textsuperscript{2}, Ji Zhang\textsuperscript{2} \\ \textbf{Fei Huang\textsuperscript{2}, Luo Si\textsuperscript{2}, Yin Zhang\textsuperscript{1}\thanks{\quad Corresponding Author: Yin Zhang.}} \\
  \textsuperscript{1}College of Computer Science and Technology, Zhejiang University, China \\
  \textsuperscript{2}DAMO Academy, Alibaba Group, China, \\
  {\tt
  \{chenqianglong,zhangyin98\}@zju.edu.cn},\\ 
  {\tt 
  \{guohai.xgh,ym119608,zj122146,f.huang,luo.si\}@alibaba-inc.com}\\
}

\begin{document}
\begin{CJK*}{UTF8}{gkai} 
\maketitle
\begin{abstract}
Existing knowledge-enhanced methods have achieved remarkable results in certain Q\&A tasks via obtaining diverse knowledge from different knowledge bases. However, limited by the properties of retrieved knowledge, they still have trouble benefiting from both the knowledge relevance and distinguishment simultaneously. To address the challenge, we propose \textbf{CPACE}, a \textbf{C}oncept-centric \textbf{P}rompt-b\textbf{A}sed \textbf{C}ontrastive \textbf{E}xplanation Generation model, which aims to convert obtained symbolic knowledge into the contrastive explanation for better distinguishing the differences among given candidates. Firstly, following previous works, we retrieve different types of symbolic knowledge with a concept-centric knowledge extraction module. After that, we generate corresponding contrastive explanation using acquired symbolic knowledge and explanation prompt as guidance for better modeling the knowledge distinguishment and interpretability. 
Finally, we regard the generated contrastive explanation as external knowledge for downstream task enhancement. 
We conduct a series of experiments on three widely-used question-answering datasets: CSQA, QASC, and OBQA. Experimental results demonstrate that with the help of generated contrastive explanation, our CPACE model achieves new SOTA on CSQA (89.8\% on the testing set, 0.9\% higher than human performance), and gains impressive improvement on QASC and OBQA (4.2\% and 3.5\%, respectively).

\end{abstract}

\section{Introduction}
In recent years, a large number of knowledge enhanced pre-trained language models (KE-PLMs)~\citep{zhang-etal-2019-ernie,weijie2019kbert,wang-etal-2021-k,wang2021KEPLER} have been proposed to improve performance on a wide variety of NLP tasks~\citep{wei2021knowledge}. 
However, the implicit knowledge learned in PLMs can not be effectively used for these knowledge-driven QA tasks, especially in commonsense question answering.
\begin{figure}[ht]
    \centering
\includegraphics[width=0.85\linewidth]{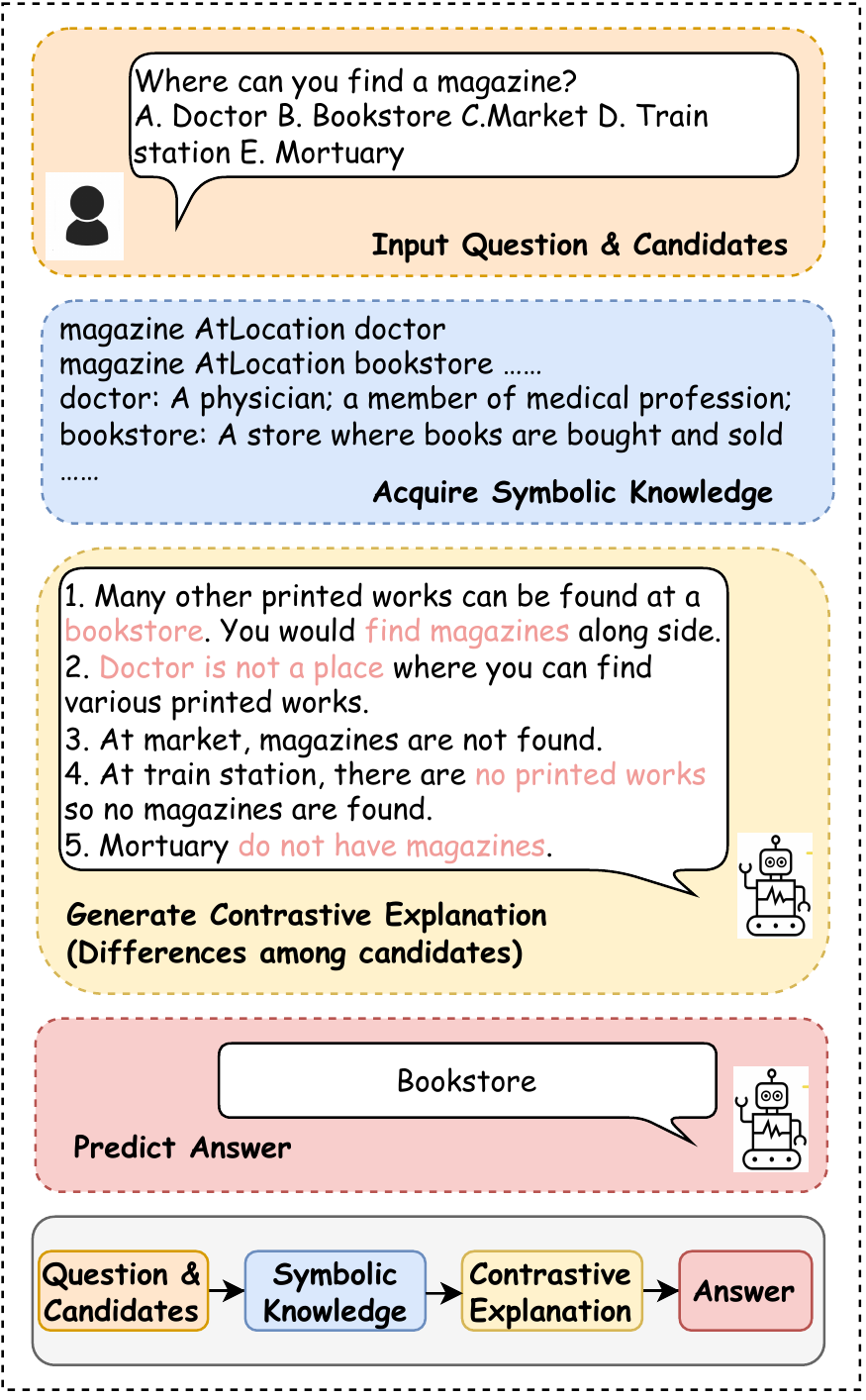}
    \caption{A motivating example for our CPACE model. To provide more distinguishing information, we can convert the acquired symbolic knowledge into contrastive explanation and use them for inference enhancement.}
    \label{fig:overview}
\end{figure}
Some works ~\citep{lv2020commonsense,wang-etal-2020-connecting,chen-etal-2020-improving,xu-etal-2021-fusing,wang-etal-2021-generative} explicitly retrieve knowledge from different knowledge sources, including WordNet~\citep{miller1995wordnet}, Wikidata~\citep{vrandevcic2014wikidata} and ConceptNet~\citep{speer2017conceptnet}, then integrate them into downstream models for Q\&A. 
These methods enjoy the ability to utilize diverse knowledge, but inevitably introduce irrelevant or even noisy knowledge which will hurt the performance of model.
Other works consider PLMs as knowledge bases~\citep{petroni-etal-2019-language,roberts-etal-2020-much,heinzerling-inui-2021-language,wang-etal-2021-generative}, which elicit potential knowledge via prompt from PLMs~\citep{paranjape-etal-2021-prompting,liu-etal-2022-generated}. These approaches can obtain relevant knowledge from PLMs, however the generated knowledge from PLMs is generally common and lacks specific and distinguishing information for enhancement. It is an important direction to explore ``\textit{how to provide discriminative information to models to help them distinguish candidates before answering ?}''.

Inspired by previous studies~\citep{chen-etal-2021-kace,paranjape-etal-2021-prompting,jacovi-etal-2021-contrastive}, contrastive explanation can provide the information to explain ``WHY A NOT B'' for given input and prediction, which naturally has distinguishing property.
As shown in Figure~\ref{fig:overview}, given question, candidates and retrieved symbolic knowledge, we generate contrastive explanation for each candidate to provide discriminative information among them for inference enhancement.
Therefore, in this paper, we propose a \textbf{C}oncept-centric \textbf{P}rompt-b\textbf{A}sed \textbf{C}ontrastive \textbf{E}xplanation generation (\textbf{CPACE}) model, a \textit{distinguish before answer} architecture, to obtain high-quality incorporated knowledge and distinguish the differences among candidates. 
Specifically, our model consists of three parts, namely \textit{symbolic knowledge acquisition} module, \textit{contrastive explanation generation} module and \textit{explanation enhanced inference} module. Firstly, given the question and candidates, we use a trained concept recognizer to detect concepts appearing in input. Then, with identified concepts, we extract diverse symbolic concept-centric knowledge from different types of knowledge bases. After that, we take the retrieved knowledge and a pre-defined explanation prompt as guidance for a fine-tuned generative pre-trained language model to generate contrastive explanation. The process of generation can filter irrelevant knowledge and convert selected symbolic knowledge into more specific and distinguishing information according to question and candidates. Finally, we use the generated contrastive explanation as external knowledge for enhancement. It is worth noting that contrastive explanation, as the final form of incorporated knowledge, not only meet distinguishing property, but also makes it easier for human to understand and is better interpretable.

The contributions are summarized as follows:
\begin{itemize}
    
    \item Based on previous exploration of contrastive explanation, we first propose a CPACE model to unify the retrieved knowledge into contrastive explanation, which can distinguish the difference among answers before prediction.
    
    \item To better adapt contrastive explanation to question answering tasks, we develop a concept-centric prompt-based generator, which can leverage concept-centric knowledge and explanation prompt as guidance.
    
    \item Our CPACE model achieves new SOTA on CSQA leaderboard~\footnote{\url{https://www.tau-nlp.sites.tau.ac.il/csqa-leaderboard}}, which surprisingly surpasses human performance. Experimental results demonstrate the generalization of our methods on QASC and OBQA datasets and the effectiveness of contrastive explanation as another type of unified knowledge form for knowledge enhancement.
\end{itemize}

\begin{figure*}
    \centering
    \includegraphics[width=0.95\linewidth]{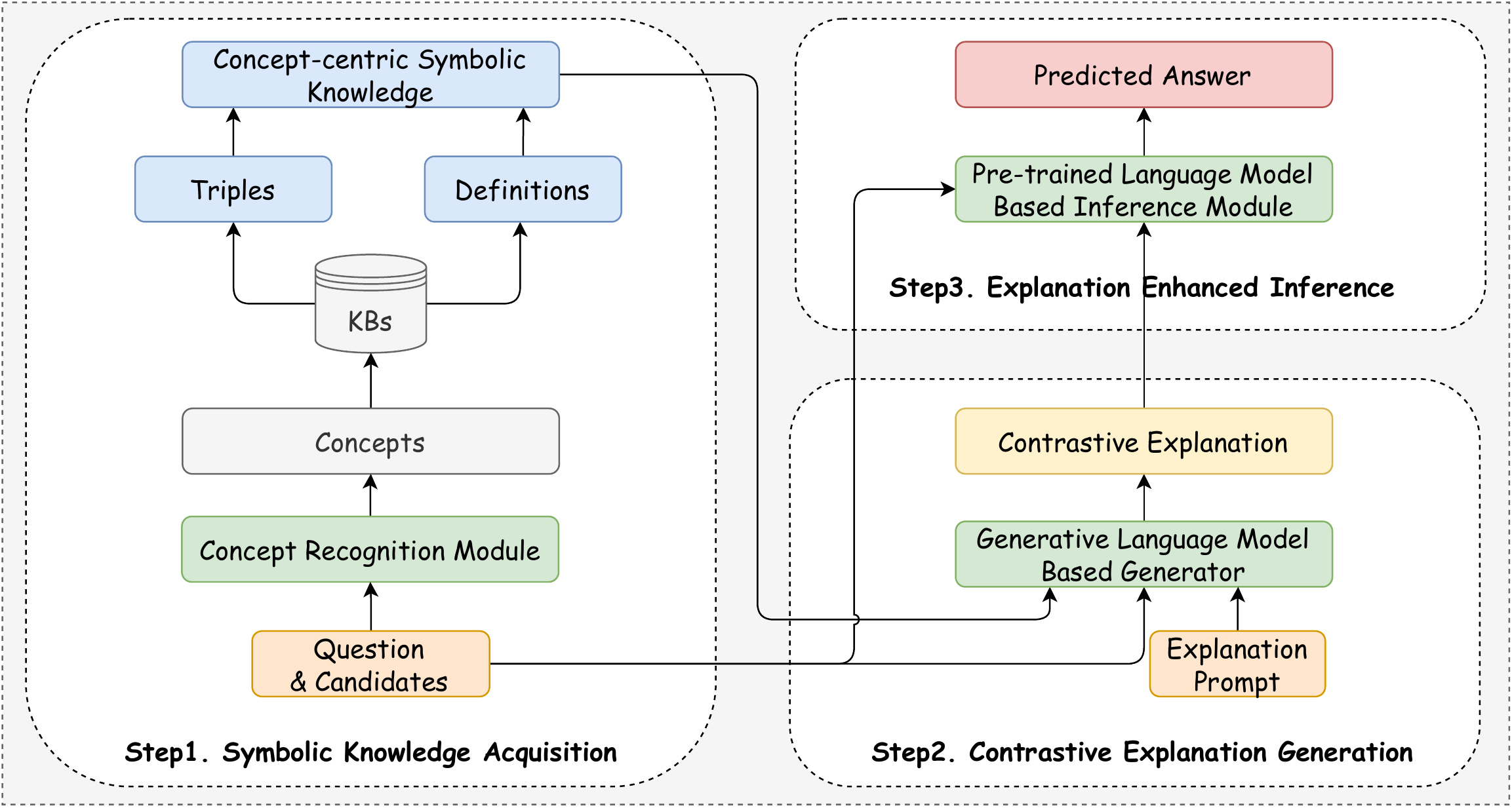}
    \caption{Architecture of our CPACE model, which consists of 1) a symbolic knowledge acquisition module, 2) a contrastive explanation generation module and 3) an explanation enhanced inference module. 
    }
    \label{fig:workflow}
\end{figure*}

\section{Task Formulation and Overall Workflow}
Here, we introduce the commonsense question answering task and the workflow of our CPACE model. 
Given a question stem $Q$, the task is to find the correct answer $a$ from a finite set of choices $A=\{a_1,a_2,...,a_n\}$. 
As shown in Figure~\ref{fig:workflow}, our approach can be divided into three steps. 
The first step is \textit{symbolic knowledge acquisition}, we build a concept recognizer 
to identify a concept set $C$ from the given question $Q$ and candidates $A$, then we take them as queries to extract diverse symbolic knowledge $K_{symbolic}$ from several knowledge bases $KBs$, as shown in Section ~\ref{sec:symbolic_knowledge_acquisition}:
\begin{equation}
\label{Concept_mining}
\begin{adjustbox}{max width=0.85\linewidth}
    \texttt{$C$ = \texttt{Recognition}($Q$, $A$)}
\end{adjustbox}
\end{equation}
\begin{equation}
\label{knowledge_extraction}
\begin{adjustbox}{max width=0.85\linewidth}
     \texttt{$K_{symbolic}$ = \texttt{Extraction}($C$, $KBs$)}
\end{adjustbox}
\end{equation}
The second step is \textit{contrastive explanation generation}, where we generate contrastive explanation $K_{ce}$ with CPACE generator, given $Q$, $A$, $K_{symbolic}$, $C$ and explanation prompt $P$, as shown in Section \ref{sec:CPACE Generator}:
\begin{equation}
\label{generation}
\begin{adjustbox}{max width=0.85\linewidth}
     \texttt{$K_{ce}$ = \texttt{Generation}($Q$, $A$, $K_{symbolic}$, $C$, $P$)}
\end{adjustbox}
\end{equation}
The final step is \textit{explanation enhanced inference}, we obtain the predicted answers $a$ from a standard inference model enhanced with $K_{ce}$, as presented in Section ~\ref{sec:Inference}:
\begin{equation}
\label{inference}
\begin{adjustbox}{max width=0.85\linewidth}
    \texttt{$a$ = \texttt{Inference}($Q$, $A$, $K_{ce}$)}
\end{adjustbox}
\end{equation}

\section{Approach}

\subsection{Symbolic Knowledge Acquisition}
\label{sec:symbolic_knowledge_acquisition}
\subsubsection{{Concept Recognition}}
\label{sec:concepts}
Considering the concepts represent the key information of examples in semantic level, some works~\cite{chen-etal-2021-kace,antognini-faltings-2021-rationalization,stowe-etal-2021-metaphor} build a connection with external knowledge through concepts. Inspired by these studies, we employ a concept recognizer to detect the concepts from given question and candidates, which can ensure the retrieved symbolic knowledge is more concept-centric and relevant to the input in external knowledge extraction. 

We first formulate concept recognition as a token-level sequence labeling task~\citep{thorne-etal-2019-generating}, where 1 indicates a concept token and 0 indicates a background token. 
For the concept recognizer, we adopt RoBERTa-large as the encoder with a CRF layer. 
We construct the input sentence \textit{S=[CLS]Q[SEP]A[SEP]}, where \textit{[SEP]} is special token to separate question and candidates. Given a sentence $S=\{t_1,t_2,...,t_n\}$, the task is to find a set of concepts $C=\{c_1,...,c_m\}$.
Limited by the scale of training corpus, we collect several similar datasets for concept recognizer training, including CommonGen~\citep{lin-etal-2020-commongen}, e-SNLI~\citep{NIPS2018_8163} and CSQA~\citep{talmor-etal-2019-commonsenseqa}, all of which contained the annotated concepts or tokens in examples. The statistics of these datasets are shown in Table \ref{tab:Statistics-1}.
While the CommonGen dataset is annotated to generate sentence with given concepts, we invert the target sentence into an input and use the given concepts as target. If there are more than 3 identified concepts in question stem, the top 3 concepts will be selected based on the score ranking mechanism for subsequent use. Otherwise, we select all identified concepts. 

\begin{figure*}
    \centering
    \includegraphics[width=0.95\linewidth]{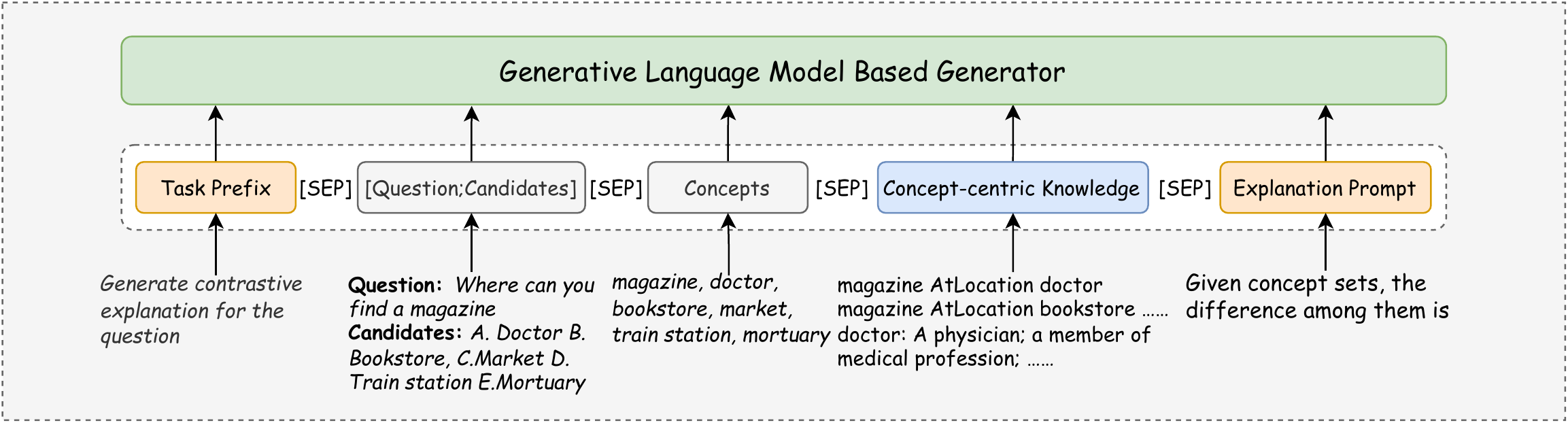}
    \caption{Details of inputs for contrastive explanation generation, where the \textit{Concepts} in the input is optional. The \textit{Explanation Prompt} we present is the best prompt used for explanation generation.
    }
    \label{fig:generator}
\end{figure*}

\subsubsection{External Knowledge Extraction}
\label{sec:extraction}
After obtaining a group of concepts, we use them as anchors to retrieve relevant external symbolic knowledge. Following previous works~\citep{chen-etal-2020-improving,xu-etal-2021-fusing}, we choose ConceptNet and Cambridge Dictionary as knowledge bases for triples and definitions extraction.
\paragraph{Triples Extraction}
To extract relationships between concepts, being similar to ~\citet{Lin-KCR}, we find the path from the question concept to the candidate concept in ConceptNet.
If there are more than one path, we choose the shortest. 
If there is no straightforward path between question concept and candidate concept, but we can find other triples in the ConceptNet with candidate concept. We define a score function and use it to compute the final score of each triples and chose the highest,
\begin{equation}
    score_j = w_j * \frac{N}{N_k}
\end{equation}
where $w_j$ denotes the weight of $j_{th}$ triple in ConceptNet, $N$ is the total number of triples related to candidate concepts, all $N$ triples related to candidate concepts are divided into multiple relation groups by clustering, and $N_k$ is the number of triples contained in $k_{th}$ relation group.

\paragraph{Definitions Extraction}
To extract definitions of concepts, following recent works~\citep{chen-etal-2020-improving, xu-etal-2021-fusing}, we obtain them from Cambridge Dictionary. For each concept, we choose its first definition entry in Dictionary as the description. When the closest matching definition entry is selected as the concept description in the dictionary, if there are multiple forms of definition entries, the priority order selected as the concept description is: the original form of the concept itself > the lemma form by Spacy~\footnote{\url{https://spacy.io/}} > base word (last word). Finally, we concatenate the triples and concept definitions as external concept-centric knowledge, 
specifically, we take \textit{Triples [SEP] Definitions [SEP]} as the \textit{Concept-centric Knowledge} for contrastive explanation generation and downstream inference.

\subsection{Contrastive Explanation Generation}
\label{sec:CPACE Generator}
In this part, we present how to generate contrastive explanations, given the question, candidates, and the retrieved knowledge, from data collection and generator training aspects. 
\paragraph{Data Collection}
Firstly, for contrastive explanation generator training, the most important thing is to collect a certain number of annotated contrastive explanation datasets.
We firstly collect some explanation-related datasets with the following principles in order: 1) whether the dataset directly contains contrastive explanations; 2) if not, can the dataset provide explanations for different candidates, i.e. positive and negative explanations; 3) if not, does the explanation of the dataset contain factual knowledge to distinguish different candidates or labels.
Therefore, we choose the training set of ECQA~\citep{aggarwal-etal-2021-explanations}, eQASC~\citep{jhamtani-clark-2020-learning} and e-SNLI~\citep{NIPS2018_8163} for generator training. The statistics of datasets are shown in Table \ref{tab:Statistics-1}.

\paragraph{Generator Training}
With the collected datasets, we train a contrastive explanation generator by fine-tuning a generative language model (GLM). In this work, we use BART-base as the backbone. In the fine-tuning stage, different from concatenating question stem and candidates in ECQA and eQASC, the hypothesis and premise sentence in e-SNLI are used as original input of GLM. The target is the explanation text.
Moreover, different from previous works only consider original questions and candidates as input for fine-tuning, we also take the concepts and external symbolic knowledge to enhance the input for the prompt-based generation. As shown in Figure~\ref{fig:generator}, the input is organized as follows: \textit{Task Prefix [SEP] [Question;Candidates] [SEP] Concepts [SEP] Concept-centric Knowledge [SEP] Explanation Prompt}, 
where \textit{Task Prefix} is ``\textit{Generate the contrastive explanation for this question}'', \textit{Concept-centric Knowledge} represents extracted symbolic knowledge (triples and definitions of concepts) shown in section ~\ref{sec:extraction}, and \textit{Explanation Prompt} are the selected discrete prompts constructed by human, which are shown in  Table~\ref{tab:prompts}. 

Different from previous work~\cite{paranjape-etal-2021-prompting} constructs Cloze prompt patterns for comparing the differences between two candidates, we consider whole contrastive explanation among all candidates and construct different discrete explanation prompts for guidance, for example, ``\textit{Given concept sets, the difference among them is }''. We use a list of templates $t_1, ..., t_p$ to generate a list of candidate explanations $e_1, ..., e_p$ for each input during fine-tuning and select the best prompt for generation, $p$ denotes the number of the templates. It is worth noting that we firstly leverage the extracted symbolic knowledge and concepts to improve the quality of generated contrastive explanation, which is ignored in ~\citet{paranjape-etal-2021-prompting}.



\subsection{Explanation Enhanced Inference}
\label{sec:Inference}
As shown in step 3 of Figure~\ref{fig:workflow}, given original question, we use the generated contrastive explanation as external knowledge to enhance the inference model, such as ALBERT and DeBERTaV3. Other types of knowledge can also be incorporated, which is optional. The objective function is defined as follows:
\begin{equation}
\begin{adjustbox}{max width=0.85\linewidth}
 $L_{ce}$ = $-\frac{1}{T}\sum^{T}\limits_{i=1}y_i log\texttt{softmax($h_i$)}$
\end{adjustbox}
\end{equation}
\begin{equation}
\begin{adjustbox}{max width=0.85\linewidth}
$\texttt{softmax($h_i$)} = {\frac{\texttt{exp($h_i$)}}{\sum_{j=1}^{n}\texttt{exp($h_j$)}}}$
\end{adjustbox}
\end{equation}
where $i$ represents the $i_{th}$ example, \texttt{$h_i$} represents the hidden state after task-specific layer (MLP), \texttt{$y_i$} represents the label of $i_{th}$ example, $T$ represents the total number of examples.


\section{Experiments}
\subsection{Datasets}
\paragraph{CSQA \& ECQA}
CommonsenseQA (CSQA) ~\cite{talmor-etal-2019-commonsenseqa} is proposed to explore the commonsense understanding ability of PLMs. To explore the interpretability of question-answering models, ECQA~\citep{aggarwal-etal-2021-explanations} is proposed with the positive and negative explanations annotated for each question in CSQA. Here, we construct the positive and negative explanations as the ground truth contrastive explanation.

\paragraph{QASC \& OBQA} 
To further validate the generalization of our CPACE model, we evaluate the effectiveness of generated contrastive explanation on QASC~\citep{khot2020qasc} and OpenbookQA (OBQA)~\citep{OpenBookQA2018}. The statistics of above datasets are shown in Table~\ref{tab:Statistics-2}. 

\subsection{Experimental Setting}
We choose BART-base ~\citep{lewis-etal-2020-bart} as the pre-trained generative language model, which is the backbone of our contrastive explanation generator. For the framework, we use Pytorch 1.11. We use the AdamW~\citep{loshchilov2018decoupled} for optimization and set the warmup fraction to 0.1, and weight decay to 0.01. Meanwhile, we set the epoch to 10. For the learning rate, we search from {1e-5, 5e-5, 1e-4} and the best batch size we choose is 32.
We set the max-length of output in the generator to 256. For the automatic evaluation, we use the ROUGE ~\citep{lin-2004-rouge} and BLEU ~\citep{papineni2002bleu} as the metric to measure the quality of generated explanation.
For the inference models, we use ALBERT-xxlarge-v2 ~\citep{Lan2020ALBERT:} and DeBERTaV3 ~\citep{he2021debertav3} as backbone respectively, which are enhanced with contrastive explanation. For each experiment, we run 5 times and report the average and we use RTX6000 with 40G memory for training and inference.

\begin{table}
    \centering
    \caption{Statistics of ECQA, eQASC, e-SNLI and CommonGen, used for CPACE generator training and concept identifier training.}
    \adjustbox{width=0.48\textwidth}{
    \begin{tabular}{|l|c|c|c|}
    \hline
        \textbf{Dataset} & \textbf{Train} & \textbf{Dev} & \textbf{Has Explanation}\\
    \hline
         ECQA & 9,741 & 1,221 & \ding{51}\\
         eQASC & 8,134 & 926 & \ding{51}\\
         e-SNLI & 549,369 & 9,843 & \ding{51}\\
         CommonGen & 67,389 & 4,018 & \ding{55} \\
    \hline
    \end{tabular}}
    \label{tab:Statistics-1}
\end{table}

\begin{table}
    \centering
    \caption{Statistics of CSQA, QASC and OBQA, used for QA task inference.}
    \adjustbox{width=0.48\textwidth}{
    \begin{tabular}{|l|c|c|c|c|}
    \hline
        \textbf{Dataset} & \textbf{Train} & \textbf{Dev} & \textbf{Test} & \textbf{Number of candidates}\\
    \hline
         CSQA & 9,741 & 1,221 & 1,140 & 5 \\
         QASC & 8,134 & 926 & 920 & 8 \\
         OBQA & 4,957 & 500 & 500 & 4 \\
    \hline
    \end{tabular}}
    \label{tab:Statistics-2}
\end{table}

\subsection{Baselines}
\paragraph{Pre-trained Language Models}
For the comparison, we choose some PLMs as baselines to validate the effectiveness of backbone encoder, including DeBERTa~\citep{he2020deberta,he2021debertav3}, ALBERT~\citep{Lan2020ALBERT:}, T5~\citep{raffel2020exploring}, UnifiedQA~\citep{khashabi-etal-2020-unifiedqa}, RoBERTa~\citep{liu2019roberta} and BERT~\citep{devlin-etal-2019-bert}. 

\paragraph{Knowledge Enhanced Methods}
In knowledge-driven Q\&A tasks, the most effective methods are external knowledge-enhanced. Here, we select representative approaches as baselines, including \textit{BERT+OMCS}~\footnote{\url{https://drive.google.com/file/d/1sGJBV38aG706EAR75F7LYwCqci9ocG9i}}, \textit{RoBERTa+MHGRN}~\citep{feng-etal-2020-scalable}, \textit{RoBERTa+AIR}~\citep{yadav-etal-2020-unsupervised}, \textit{TeGBERT}~\footnote{\url{https://docs.google.com/document/d/1JdkIxKr4wehfHeHrcZrpolORHtjXgF-sJSK6e_D0m54/}},
\textit{ALBERT+KD}~\footnote{\url{https://docs.google.com/document/d/1aK4aE86H4LLCZ4-ZGuqOvVXFNybYuRnfry6kM5eqyTY}},
\textit{ALBERT+KCR}~\citep{Lin-KCR}, \textit{ALBERT+Headhunter}~\citep{li-etal-2021-winnowing-knowledge}, \textit{ALBERT+PathGenerator}~\citep{wang-etal-2020-connecting}, \textit{ALBERT+HGN}~\citep{yan-etal-2021-learning}, \textit{ALBERT+DESC-KCR}~\citep{xu-etal-2021-fusing}, \textit{GenMC}~\citep{huang-etal-2022-clues}, \textit{QA-GNN}~\cite{yasunaga2021qagnn}, and \textit{KEAR}~\citep{xu2021human}. 
More details of these baselines are shown in Appendix ~\ref{baselines}.

\subsection{Main Results}

\begin{table}[h]
    \centering
    \caption{
    Results on CSQA test set from the leaderboard. All references can be found in this document\protect\footnotemark.}
    \adjustbox{width=0.48\textwidth}{
    \begin{tabular}{|l|c|c|}
    \hline
        \textbf{Model} & \textbf{Single} & \textbf{Ensemble}\\
    \hline
    \multicolumn{3}{|c|}{\textbf{Pre-trained Language Model Only}}\\
    \hline
    BERT & 56.7 & -\\
    RoBERTa & 72.1 & 72.5 \\
    ALBERT & 73.5 & 76.5\\
    T5 & 78.1 & - \\
    UnifiedQA & 79.1 & - \\
    DeBERTa & - & 79.6 \\
    \hline
    \multicolumn{3}{|c|}{\textbf{PLM + Symbolic Knowledge Retrieval}}\\
    \hline
        BERT + OMCS & 62.5 & - \\
        RoBERTa + MHGRN & 75.4 & 76.5\\
        QA-GNN & 76.1 & - \\
        TeGBERT & 76.8 & - \\
        ALBERT + Headhunter & 78.4 & - \\
        ALBERT + KCR & 79.5 & - \\
        ALBERT + KD & 80.3 & 80.9\\
        ALBERT + DESC-KCR & 80.7 & 83.3 \\
    \hline
    \multicolumn{3}{|c|}{\textbf{PLM + Generated Knowledge}}\\
    \hline
        GenMC & 72.6 & - \\
        ALBERT + PathGenerator & 75.6 & 78.2 \\
        ALBERT + HGN & 77.3 & 80.0\\
    \hline
    \multicolumn{3}{|c|}{\textbf{Beyond Human Level}}\\
    \hline
        KEAR & 86.1 & 89.4\\
        CPACE & \textbf{87.4} & \textbf{89.8} \\
    \hline
        Human Performance & - & 88.9\\
    \hline
    \end{tabular}}
    \label{tab:leaderboard_results}
\end{table}
\footnotetext{\url{https://docs.google.com/spreadsheets/d/1B3oKAzzG6kxK6cWCckhsVt4LmAtxrzoj89vBonBkeJw/}}


As shown in Table~\ref{tab:leaderboard_results},
we divide existing methods on CSQA into four parts: 1) \textbf{Pre-trained Language Model Only}, 2) \textbf{PLM + Symbolic Knowledge Retrieval},  3) \textbf{PLM + Generated Knowledge}, and 4) \textbf{Beyond Human Level}.
Compared with all of baselines, our CPACE model achieve the best performance on CSQA. 

Specifically, in part 1 of Table~\ref{tab:leaderboard_results}, experimental results demonstrate the selection of pre-trained language models (PLM) is important for commonsense question answering. PLM with optimal pre-training tasks and large parameters achieves better results. While RoBERTa-large only achieves 72.5\%, DeBERTa obtains 79.6\% on CSQA, which adopts disentangled attention for decoding enhancement and has 1.5B parameters. In part 2 of Table~\ref{tab:leaderboard_results}, incorporating triples and concept definitions helps a lot to improve the performance of PLMs on CSQA.
Compared with ALBERT, ALBERT+DESC-KCR achieves 83.3\% on CSQA, which gains 7.8\% improvement. 
Meanwhile, other works attempt to generate triples or relationships with PLMs, as shown in part 3 of Table~\ref{tab:leaderboard_results}. While ALBERT + PathGenerator only achieves 75.6\% via dynamically generating structured evidence, our CPACE model achieves 87.4\% in single model setting via generating contrastive explanations.
Furthermore, while KEAR leverages external knowledge and retrieved training example for knowledge enhancement, our CPACE model outperforms KEAR and achieves first place on CSQA leaderboard.

Overall, while ALBERT achieves 73.5\% on CSQA test set, existing knowledge-enhanced methods achieve 3.8\%-7.2\% improvement and our CPACE model improves over 13.9\%. It indicates the generated contrastive explanation can be another efficient way for knowledge enhancement instead of retrieving triples, definitions, and training examples. 
It is noted that while KEAR joints human party via  extra training examples retrieval and using over 39 models for ensemble, we only use 5 models for ensemble and propose a contrastive explanation generator, which is easier to follow.

\subsection{Generalization of CPACE}
\label{sec:Generalization}
To further measure the generality of CPACE, we evaluate our model on QASC and OBQA datasets. 
As shown in Table~\ref{tab:eQASC_eOBQA}, we select some representative baselines for comparison, including UnifiedQA, RoBERTa+AIR and GenMC. Although ALBERT only gets 71.8\% and 72.5\% on QASC and OBQA, ALBERT + KD achieves 80.3\% and 83.2\% respectively, which only retrieves symbolic knowledge from KBs. With our CPACE model, we can further improve by 3.4\% and 2.9\%, respectively. The experimental results show that our CPACE model can be used not only for commonsense question answering but also for other open-domain Q\&A. Meanwhile, we present the case study in Appendix~\ref{case_study}.

\begin{table}
    \centering
    \caption{Results on development set of QASC and OBQA, demonstrating the generalization of CPACE.}
    \adjustbox{width=0.38\textwidth}{
    \begin{tabular}{|l|c|c|}
    \hline
        \textbf{Model} &  \textbf{QASC} & \textbf{OBQA}\\
    \hline
        BERT & 68.4 &  64.1 \\
        UnifiedQA & 66.6 & 70.5 \\
        GenMC & 67.6 & 71.6 \\
        ALBERT & 71.8 & 72.5\\
        ALBERT + KD & 80.3 & 83.2 \\
        RoBERTa + AIR & 81.4 & 81.7 \\
    \hline
        CPACE & \textbf{83.7} & \textbf{86.1}\\
    \hline
    \end{tabular}}
    \label{tab:eQASC_eOBQA}
\end{table}

\begin{table}[t]
    \centering
    \caption{Ablation study of generator on development set of CSQA. We adopt ALBERT as the inference model.}
    \begin{adjustbox}{width=0.48\textwidth}
    \begin{tabular}{|l|c|}
    \hline
         \textbf{Model} & \textbf{Dev Accuracy} \\
    \hline
         BART & 78.3\\
         BART + Concept & 79.1\\
         BART + Explanation Prompt & 82.4 \\
         BART + Concept-centric Knowledge & 83.5\\
         BART + All & \textbf{85.2}\\
    \hline
    \end{tabular}
    \end{adjustbox}
    \label{tab:generator_ablation_study}
\end{table}

\subsection{Ablation Study}
\paragraph{Analysis of Contrastive Explanation Generator}
As shown in Table~\ref{tab:generator_ablation_study}, we use BART-base as the backbone to evaluate the effectiveness of concepts, prefix prompt, and retrieved concept-centric knowledge (triples and definitions of concepts) in the generator.
Only with the fine-tuned BART-base as the generator, the generated explanation enhanced inference model can achieve 78.3\% on CSQA development set. 
Since concepts represent the key information of a given sentence, with identified concepts, the generator can get some benefits. When taking concepts as enhanced input, we can obtain 0.8\% 
improvement. When taking explanation prompt as a formal constraint, we get an improvement of 4.1\% 
, which fully shows the necessity of contrastive explanation prompt as constraint. Meanwhile, enhanced with the external concept-centric knowledge, we can gain 5.2\% 
improvement, which indicates concept-centric knowledge is equally important in contrastive explanation generation. 
Finally, with the incorporated of above three kinds of knowledge, the inference model can be improved by 6.9\%.


\paragraph{Analysis of Inference Encoder}
In this part, we use ALBERT-xxlarge-v2 and DeBERTaV3 as the inference encoder. As shown in Table~\ref{tab:inference_ablation_study}, ALBERT achieves 73.8\% on CSQA, the DeBERTa achieves 84.6\%, which indicates a better inference backbone is of importance in the downstream task. Then, we take the concept, retrieved concept-centric knowledge and generated contrastive explanation as different types of extra knowledge to enhance the inference model, respectively. While concept can only bring about 1.5\% 
and 0.2\% 
improvement for ALBERT and DeBERTa, we can get 10.4\% 
and 0.5\% 
improvement through triples and concept definitions, respectively. With the generated contrastive explanation, we can get a great improvement, which is 11.4\% 
and 3.3\% 
respectively. It demonstrates that generated contrastive explanation is much more effective than retrieved symbolic knowledge. 
Compared with adding ground-truth contrastive explanation, which achieves 11.7\%
and 9.2\% improvement respectively, there is still some room for improvement.

\begin{table}
    \centering
    \caption{Ablation study of inference encoder on development set of CSQA. We enhance downstream models with different types of knowledge.}
    \adjustbox{width=0.48\textwidth}{
    \begin{tabular}{|l|c|}
    \hline
         \textbf{Model} & \textbf{Dev Accuracy}\\
    \hline
         ALBERT & 73.8 \\
         ALBERT + Concept & 75.3 \\
         ALBERT + Concept-centric Knowledge & 84.2 \\
         ALBERT + Contrastive Explanation & 85.2\\
         ALBERT + All & \textbf{88.4}\\
    \hline
         DeBERTaV3 & 84.6\\
         DeBERTaV3 + Concept & 84.8\\
         DeBERTaV3 + Concept-centric Knowledge & 85.1\\
         DeBERTaV3 + Contrastive Explanation & 87.9\\
         DeBERTaV3 + All & \textbf{91.7}\\
    \hline
        ALBERT + Ground-truth Explanation & 96.9 \\
        DeBERTaV3 + Ground-truth Explanation & \textbf{97.1}\\
    \hline
    \end{tabular}}
    \label{tab:inference_ablation_study}
\end{table}


\begin{table*}[t]
    \centering
    \caption{Human evaluation of generated contrastive explanation on development set of CSQA.}
    \adjustbox{width=1\textwidth}{
    \begin{tabular}{|l|c|c|c|c|}
    \hline
       \textbf{Model}  & \textbf{Relevant} & \textbf{Factual} & \textbf{Distinguishing} & \textbf{Grammatical}\\
    \hline
        BART & 68.2 & 47.0 & 26.0 & 83.2 \\
        BART + Concept & 71.0 & 48.2 & 28.2 & 83.6\\
        BART + Explanation Prompt & 72.4 & 48.6 & 29.1 & 83.7\\
        BART + Concept-centric Knowledge & \multirow{1}{*}{75.3} & \multirow{1}{*}{52.2} & \multirow{1}{*}{50.1} & \multirow{1}{*}{84.2} \\ 
        BART + All & \textbf{80.3} & \textbf{54.6} & \textbf{53.4} & \textbf{87.5}\\
    \hline
    \end{tabular}
    }
    \label{tab:human_evaluation}
\end{table*}

\subsection{Evaluation of Contrastive Explanation}


As shown in Table~\ref{tab:human_evaluation}, following~\citet{shwartz-etal-2020-unsupervised}, we present the human evaluation of generated contrastive explanation in four aspects, including 1) \textbf{Relevant}, whether the generated explanation is relevant to current example, 2) \textbf{Factual}, if the explanation contains factual evidence, 3) \textbf{Distinguishing}, if the explanation can provide distinguishing information to improve inference, and 4) \textbf{Grammatical}, whether the generated explanation is grammatical. 

We sample 100 explanations from generated contrastive explanation on CSQA and evaluate the score of the generated explanation from above aspects. We use five students as annotators and report the average. As we can see, taking the given example as input of BART-base, we only get high grammatical score but with low distinguishing score. When we take concept and explanation prompt for enhancement, it can slightly improve the the relevance and distinguishment. 
Enhanced with concept-centric knowledge, it can improve the distinguishing score over 20\%, which is much more helpful. Furthermore, we can use all the above knowledge to get the best performance.
Besides, we demonstrate automatic evaluation of contrastive explanation, which is shown in Appendix ~\ref{evaluation}.

\section{Related Work}
\subsection{Knowledge Enhanced Methods}
To alleviate the knowledge insufficiency problem, many knowledge-enhanced works have been proposed~\citep{ijcai2022p0567, 10.1145/3503161.3548406, weijie2019kbert,wang2021KEPLER,wang-etal-2021-k,chen-etal-2020-improving,zhang2021dkplm,Chen2022RethinkingTV,10.1145/3459637.3481930}, which can be roughly categorized into explicit symbolic knowledge retrieval based and implicit knowledge generation based. In the former works, researchers~\citep{lv2020commonsense,chen-etal-2020-improving,xu-etal-2021-fusing} mainly focus on acquiring relevant knowledge from different knowledge bases, including ConceptNet ~\citep{speer2017conceptnet}, Wikipedia, and dictionaries. 
These methods enjoy the benefits of diverse knowledge but inevitably introduce irrelevant or even noisy knowledge. In the latter works, attempts ~\citep{petroni-etal-2019-language,gao-etal-2021-making,schick-schutze-2021-just,zhong-etal-2021-factual,ijcai2022p0567} have been made to explore the possibility of using pre-trained language models~\citep{devlin-etal-2019-bert,peters-etal-2018-deep} as a knowledge base.
While ~\citet{petroni-etal-2019-language} first regard PLMs as knowledge bases, other works~\citep{gao-etal-2021-making,schick-schutze-2021-just,zhong-etal-2021-factual} use different prompt-based methods to elicit potential knowledge from PLMs. 
However, limited by the pre-training corpus, the generated knowledge from PLMs lacks specific information. To takes both advantages of symbolic knowledge retrieval and knowledge generation, we propose the \textit{distinguish before answer} framework to generate contrastive explanation.


\subsection{Contrastive Explanation}
Contrastive explanations clarify why an event occurred in contrast to another, which are inherently intuitive to humans to both produce and comprehend~\cite{jacovi-etal-2021-contrastive}. Compared to other explanations, ~\citet{miller2019explanation} first suggests contrastive explanations are more effective in human learning. While ~\citet{liang-etal-2020-alice} first leverage expert annotated contrastive explanations for active learning to improve data efficiency, ~\citet{jacovi-etal-2021-contrastive} propose a method to produce contrastive explanations automatically in the latent space via input token/span for 3-label classification. Meanwhile,  ~\citet{chen-etal-2021-kace} generate contrastive explanation with counterfactual examples for natural language inference. Different from ~\citet{paranjape-etal-2021-prompting} uses human templates to prompt PLMs to generate contrastive explanations, we focus on generating contrastive explanation with retrieved symbolic knowledge to distinguish the candidates before prediction, which is ignored in previous works. ~\citet{liu-etal-2022-generated} and \citet{huang-etal-2022-clues} are the concurrent works, focusing on improving question answering with generated knowledge and clues.

\section{Conclusion}
In this paper, we propose a CPACE model, which unify the retrieved knowledge into contrastive explanation, to provide more discriminative information for model enhancement. We firstly consider concept-centric knowledge and explanation prompt as guidance for contrastive explanation generation. Our CPACE model achieves a new SOTA on CSQA leaderboard, which surprisingly surpasses human performance. In addition, we verify the effectiveness and generalization of CPACE on other datasets. In the future, we will explore a unified contrastive explanation generation framework for NLP tasks.



\section{Limitations}
Limited by the scale of annotated contrastive explanation corpus, our CPACE model is only fine-tuned on approximate datasets selected with some designed principles. The performance of our method can be further improved with sufficient high-quality contrastive explanation annotated datasets over more NLP tasks. 
Moreover, in this paper, we mainly explore the effectiveness of the CPACE model for multiple-choice commonsense question-answering tasks, which is our goal, while previous retrieved-augmented methods cannot provide highly relevant knowledge or context for reasoning. 
Due to the fact that the contrastive explanation is designed to provide distinguishing information among given options $[a_1, a_2,…,a_n]$ or labels, there are no given candidates or labels in generative commonsense question-answering tasks, therefore, our CPACE model cannot directly fit to other generative QA benchmark datasets. However, in our work, we provide some insights for future exploration, that is, generating question-specific distinguishing knowledge with a contrastive explanation generator can improve the performance and interpretation of current reasoning models. 
Meanwhile, although we validate the generalization of CPACE on other QA tasks, including QASC and OBQA, the effectiveness of our model in other NLP tasks requiring contrastive knowledge enhancement, such as open domain dialogue, needs to be further explored. In the future, following the CPACE model, we will explore a unified contrastive explanation generation framework for the generative commonsense question answering tasks via generating the chain-of-thoughts with a large generative language model-based generator, such as InstructGPT~\cite{Ouyang2022TrainingLM}, BLOOM~\cite{Scao2022BLOOMA1} etc., or generating top-N possible candidates and ranking them with distinguishing knowledge, which is beyond the scope of this paper to explore and is also our future work.

\section*{Acknowledgments}
We thank the anonymous reviewers for their helpful comments on this paper. This work was supported by the NSFC under Grant No.~62072399, Zhejiang Provincial Natural Science Foundation of China under Grant No.~LZ23F020009, Chinese Knowledge Center for Engineering Sciences and Technology, MoE Engineering Research Center of Digital Library, Alibaba Group, Alibaba-Zhejiang University Joint Research Institute of Frontier Technologies, and the Fundamental Research Funds for the Central Universities.
\bibliography{anthology,custom}
\clearpage
\newpage
\appendix

\section{Details of Baselines}
\label{settings}
\subsection{Baselines}
\label{baselines}
\paragraph{BERT} BERT~\citep{devlin-etal-2019-bert} is the traditional pre-trained language model with mask language modeling and next sentence prediction pre-training tasks, which is used in most NLP tasks.
\paragraph{RoBERTa} RoBERTa~\citep{liu2019roberta} further optimizes BERT via pre-training on more corpus and removing next sentence prediction task. 
\paragraph{ALBERT} ALBERT~\citep{Lan2020ALBERT:} is proposed to lower memory consumption and increase the training speed of BERT and focus on modeling inter-sentence coherence via a self-supervised loss, which is also widely used as backbone.
\paragraph{T5} To explore the landscape of transfer learning techniques for NLP, T5~\citep{raffel2020exploring} introduces a unified framework that converts all text-based language problems into a text-to-text format and achieves new SOTA on many benchmarks.
\paragraph{UnifiedQA} UnifiedQA~\citep{khashabi-etal-2020-unifiedqa} is proposed to cross the boundaries among QA tasks via a single pre-trained QA model.
\paragraph{DeBERTa} To improve the BERT and RoBERTa models, ~\citet{he2020deberta} propose DeBERTa with disentangled attention mechanism and an enhanced mask decoder. And they~\cite{he2021debertav3} further optimize DeBERTa via ELECTRA-style~\citep{clark2020electra} pre-training with gradient-disentangled embedding sharing.

\begin{table*}[t]
    \centering
    \caption{Evaluation of generated contrastive explanation on CSQA with BLEU and ROUGE metrics.}
    \adjustbox{width=1\textwidth}{
    \begin{tabular}{|l|c|c|c|c|c|}
    \hline
        \textbf{Model} & \textbf{ROUGE-1} & \textbf{ROUGE-2} & \textbf{ROUGE-L} & \textbf{ROUGE-SUM} & \textbf{BLEU-1} \\
    \hline
    BART & 25.7 & 10.7 & 20.9 & 23.1 & 53.6  \\
    BART + Concept & 27.1 & 13.4 & 21.8 & 24.7 & \textbf{60.2} \\
    BART + Prefix Prompt & 28.7 & 14.6 & 22.3 & 25.1 & 59.7 \\
    BART + Concept-centric Knowledge& 48.6 & 23.7 & 34.0 & 43.1 & 45.1 \\
    BART + All & \textbf{50.6} & \textbf{24.6} & \textbf{35.8} & \textbf{45.5} & 47.1 \\
    \hline
    \end{tabular}}
    \label{tab:BLEU_ROUGE_eval}
\end{table*}

\paragraph{BERT+OMCS} BERT+OMCS finetunes BERT-large “whole word masking” model on the Open Mind Common Sense (OMCS) corpus used for creating ConceptNet.
\paragraph{TeGBERT} TeGBERT is a multi-modal learning method for commonsense reasoning, where paths are searched from a given question and choice with ConceptNet with triple scoring and triples are pre-trained with kg2vec such as transE~\citep{NIPS2013_1cecc7a7}.

\paragraph{RoBERTa+MHGRN} ~\citet{feng-etal-2020-scalable} propose RoBERTa+MHGRN to equip pre-trained language models with a multi-hop graph relation network (MHGRN) module, which performs multi-hop, multi-relational reasoning over sub-graphs extracted from external knowledge graphs.
\paragraph{RoBERTa+AIR} RoBERTa+AIR~\cite{yadav-etal-2020-unsupervised} is a method with alignment-based iterative retriever, which retrieves high-quality evidence sentences from unstructured knowledge bases and achieves new SOTA on QASC.
\paragraph{QA-GNN} ~\citet{yasunaga2021qagnn} proposed QA-GNN to identify relevant knowledge from large KGs, and perform joint reasoning over the QA context and KG via relevance scoring and joint reasoning.
\paragraph{GenMC} ~\citet{huang-etal-2022-clues} propose a generation-enhanced multiple-choice question answering (MCQA) model, GenMC, which generates a clue from the question and then leverages the clue to enhance a reader for MCQA. It outperforms text-to-text models on multiple MCQA datasets.
\paragraph{ALBERT+Headhunter} ~\citet{li-etal-2021-winnowing-knowledge} utilizes a self-attention module to re-distribute the importance of knowledge for common-sense reasoning, where top k commonsense knowledge are extracted from OMCS and they employ a Self-Attention module to interact with each triple representation.
\paragraph{ALBERT+KCR} ~\citet{Lin-KCR} propose a knowledge base method ALBERT+KCR to enhance text encoder, where they extract relevant triples from ConceptNet.
\paragraph{ALBERT+KD} ALBERT+KD combines ConceptNet and dictionary definitions for inference, where they use python's networkx library to frame ConceptNet and use the Oxford dictionary to extract the definitions of concepts.

\paragraph{ALBERT+DESC-KCR}
~\citet{xu-etal-2021-fusing} employ external entity descriptions to provide contextual information for knowledge understanding and retrieve descriptions of related concepts from Wiktionary and feed them as additional input to pre-trained language models.

\paragraph{ALBERT+PathGenerator}
~\citet{wang-etal-2020-connecting} augment a general commonsense QA framework with a knowledgeable path generator, where the generator learns to connect a pair of entities in text with a dynamic and multi-hop relational path.
\paragraph{ALBERT+HGN}
~\citet{yan-etal-2021-learning} propose Hybrid Graph Network to jointly contextualize extracted and
generated knowledge by reasoning over both
within a unified graph structure.

\paragraph{KEAR}
~\citet{xu2021human} propose KEAR to retrieve labeled examples from several question answering datasets and augment them as external knowledge for inference.


\section{Automatic Evaluation of Contrastive Explanation }
\label{evaluation}

As shown in Table~\ref{tab:BLEU_ROUGE_eval}, we also present the BLEU and ROUGE results of contrastive explanation generated with different types of knowledge. While the BLEU metric focuses on the precision of text, the ROUGE metric mainly evaluates the recall performance of generated text, which denotes the text can provide more relevant contextual information for given questions. 
Since we use the generated text as external knowledge for inference enhancement, the recall performance in the generation is much more important. 
With concepts and prompt constrained, we obtain better generated explanation text in the BLEU metric, while we can acquire better generated explanation text in the ROUGE metric with the external concept-centric symbolic knowledge (triples and definitions). 

\begin{table*}[]
    \centering
    \caption{Prompts we constructed for CPACE generator.}
    \begin{tabular}{|l|c|}
    \hline
         \textbf{Prompt Patterns} & \textbf{Task} \\
    \hline
         Given concept sets [$OPT_1, ..., OPT_n$], the difference among them is ...  & \multirow{7}{*}{CSQA QASC OBQA} \\ 
         Given [$OPT_1, ..., OPT_n$], while [$OPT_1$] can ..., ..., [$OPT_n$] can not ... & \\
         The main difference among the concepts [$OPT_1, ..., OPT_n$] is ... & \\
         Given concepts, [$OPT_1$] can, but [$OPT_2, ..., OPT_n$] can not ... & \\
         Given concepts, while [$OPT_1$] can, [$OPT_2, ..., OPT_n$] can not ... & \\
         Given concepts, [$OPT_1$] can not, however [$OPT_2, ..., OPT_n$] can ... & \\
         Given concepts, while [$OPT_1$] can not, however [$OPT_2, ..., OPT_n$] can ... & \\
         
    \hline
    \end{tabular}
    
    \label{tab:prompts}
\end{table*}

\begin{table*}[t]
    \centering
    \caption{Case study of CAPCE generator on CSQA dev set.}
    \adjustbox{width=0.98\textwidth}{
    \begin{tabular}{|l|l|}
    \hline
    \multirow{2}{*}{\textbf{Input Example}} & Where can you find a magazine. \\ & A. doctor B.bookstore, C.market D.train station E.mortuary\\
    \hline
    \textbf{Labels} & B \\
    \hline
    \textbf{Predicted Labels} & B \\
    \hline
        \textbf{Step1.1: Identified Concepts} &  magazine, doctor, bookstore, market, train station, mortuary \\
    \hline
        \multirow{4}{*}{\textbf{Step1.2: Triples from ConceptNet}} & magazine AtLocation doctor \\ & magazine AtLocation bookstore\\ & magazine AtLocation market \\ & magazine AtLocation train station\\
    \hline
        \multirow{6}{*}{\textbf{Step1.2: Concept Description}}  & \textbf{doctor}: A physician; a member of medical profession; \\ &  one who is trained and licensed to heal the sick or injured \\ \multirow{4}{*}{\textbf{from Dictionary}}& \textbf{bookstore}: A store where books are bought and sold\\  & \textbf{market}: A gathering of people for the purchase and sale \\ &  of merchandise at a set time. \\ & \textbf{train station}: A place where trains stop for passengers \\ & to embark  and disembark. \\ &  \textbf{mortuary}: of or relating to death or a funeral; funeral;\\
    \hline
    \multirow{6}{*}{\textbf{Step2: Generated Contrastive}} & A store is a place where people can find magazines along with \\ & many other printed works. \\ \multirow{4}{*}{\textbf{Explanation only with BART-base}}& Doctor is a physician. \\ \textbf{}& You can buy something at market. \\ & At train station, you can take a train. \\ & Mortuary can be found in funeral.\\
    \hline
    
   \multirow{6}{*}{\textbf{Step2: Generated Contrastive }} & Many other printed works can be found at a bookstore. \\ & You would find magazines along side. \\ \multirow{4}{*}{\textbf{Explanation with CPACE}}& Doctor is not a place where you can find various printed works. \\ \textbf{}& At market, magazines are not found. \\ & At train station, there are no printed works so no magazines  \\ & are found. Mortuary do not have magazines.\\
    \hline
    \multirow{6}{*}{\textbf{Golden Contrastive Explanation}} & Bookstores have a variety of reading material including \\ & books, magazines, novels, etc. \\ & The doctor is not a place. Market sells various items one of \\ & which is printed works. \\ &  Train stations do not have various printed works. \\ & Mortuary has dead bodies.\\
    \hline
    \end{tabular}
    }
    \label{tab:case_study}
\end{table*}

\section{Case Study}
\label{case_study}
As shown in Table~\ref{tab:case_study}, we present a case study of our model. Given a question \textit{Where can you find a magazine} and a set of candidates \textit{\{A: doctor, B: bookstore, C: market, D: train, E: mortuary\}}, the true answer is \textit{B:bookstore}. We first identify the concepts from the input example, including concepts in question stem and answer candidates. Then, we can extract the triples from ConceptNet. As shown in Table~\ref{tab:case_study}, both four candidates have same relation with \textit{magazine}, which can not further filter the true answer. 
With adding the concepts descriptions, we can further distinguish the candidates with same relations. 
However, the description of concepts is not clear enough for explanation, compared with the annotated contrastive explanation. With our CPACE generator, we can obtain the generated contrastive explanation with concepts, symbolic knowledge and prompt enhanced, as Table ~\ref{tab:case_study} shown. 
Compared with the extracted triples and concept descriptions, the generated contrastive explanation is much easier to understand for user, while only with BART-base, we can only obtain candidates related explanation without considering question concepts. As shown in Table ~\ref{tab:case_study}, the concepts and symbolic knowledge can help the generative language model concentrated on the key difference of candidates according to question concepts. 
Meanwhile, with the generated contrastive explanation for enhancement, we can infer the predicted answer is \textit{bookstore}.


\end{CJK*}
\end{document}